\documentstyle[cogsci-1,epsf]{article}
\title{Generating Coherent Messages in Real-time
Decision Support:\\
Exploiting Discourse Theory for Discourse 
Practice}
\author{{\large \bf Sandra Carberry} \\
Department of Computer Science \\ University of Delaware \\ Newark, Delaware 19716 \\{\tt carberry@cis.udel.edu} \And
{\large \bf Terrence Harvey} \\
Department of Computer Science \\  University of Delaware\\ Newark, Delaware 19716 \\{\tt harvey@cis.udel.edu}}

\begin{document}
\nocite{Webber92,Gertner96,Hovy91a,Moore93b,Wolz90,Buchanan95,Binsted95,Hirst96,Wanner96,mannthompson87,McKeown85,Grosz77a,Zukerman95a}

\bibliographystyle{alpha}

\maketitle
\vspace{-.7in}
\begin{abstract}
This paper presents a message planner, TraumaGEN, that draws on
rhetorical structure
and discourse theory to address the problem of
producing integrated messages from individual
critiques, each of which is designed to achieve its own communicative
goal.  TraumaGEN takes into account the purpose of the messages, the
situation in which the messages will be received, and the social role
of the system. 
\end{abstract}

\section{Introduction}

The  generation of multisentential discourse
 has focused on  generating text
that accomplishes one particular rhetorical goal, such as
describing a physical device.
In contrast, to deliver real-time decision support in trauma
management, our text generation system must be able to take an
arbitrary and often inter-related set of communicative goals 
and produce a message that realizes the entire set in as concise and
coherent a manner as possible.
This paper presents our implemented system, TraumaGEN, that addresses
this problem.  It describes the strategies that we have adopted to produce
coherent integrated messages and discusses how our system takes into account
factors such as the social role of the system.

\section{Real-Time Decision-Support System}

TraumAID (Webber, Rymon, and Clarke, 1992)
is a decision support system for addressing the
initial definitive management of multiple trauma.   
TraumaTIQ (Gertner and Webber, 1996) is a  module
that compares
a physician's plan for managing patient care with TraumAID's own
management plan and critiques significant differences between them.
To hypothesize the physician's plan, TraumaTIQ
first  chains to identify  possible explanations for an action; it
then evaluates these possible explanations on the basis of relevance
to TraumAID's current management plan and evidence provided by the
physician's other actions.  
Once the best explanation(s) have been incorporated
into the system's model of the physician's plan, 
TraumaTIQ identifies differences between that
plan and TraumAID's current management plan and  notifies the physician of
those discrepancies that could seriously impact patient care.
These include actions that have been ordered but are not yet justified,
procedures that are suboptimal, scheduling errors, and omitted actions.
TraumaTIQ's critiques are conveyed using natural language sentences
generated by filling in sentence schemata. 

The
problem we address here is that, while in isolation each of TraumaTIQ's
critiques may effectively warn a physician about a problem in their plan,
in most cases when TraumaTIQ finds the physician's
plan deficient, several problems  are detected and thus multiple
critiques are produced.  We found that there was informational
overlap among the critiques, that some critiques detracted from
other ones, and that some critiques would make more sense if they took
explicit account of those appearing earlier.
Thus a text planner was needed to generate coherent and concise integrated
messages that satisfy the multiple goals of the individual critiques.

\section{Text Planning for Multiple Goals}

The system
we built to solve this problem, TraumaGEN, is presented
with several communicative goals and a means for achieving each goal
in isolation.  It uses a set of transformational rules to
transform these into coherent message units
that achieve the overall set of  goals.
These message units are then translated into natural language using
sentence schemata.

\begin{figure*}
\begin{center}
\begin{tabular}{p{6.5in}}
\multicolumn{1}{l}{\underline{Individual critiques produced by TraumaTIQ:}}
\\
* Caution: check for medication allergies and order pulmonary care
immediately to treat the left pulmonary parenchymal injury.
\\
* Caution: check for medication allergies and order pulmonary care
immediately to treat the compound rib fracture of the left chest.
\\
* Caution: check for medication allergies and do a laparotomy 
immediately to treat the intra-abdominal injury.
\\
* Caution: do a laparotomy and repair the left diaphragm immediately
to treat the lacerated left diaphragm.
\\
* Consider checking for medication allergies now to treat a possible
GI tract injury.
\end{tabular}
\end{center}
\begin{center}
\begin{tabular}{p{6.5in}}
\underline{Merged message:}
\\
Caution: check for medication allergies as part of treating the
left pulmonary parenchymal injury, treating the compound rib
fracture of the left chest, treating the intra-abdominal injury,
and treating a possible GI tract injury.  Then order pulmonary
care to complete treating the left pulmonary parenchymal
injury and treating the compound rib fracture of the left
chest, and do a laparotomy to complete treating the intra-abdominal
injury.
\end{tabular}
\end{center}
\caption{Original Critiques and a Merged Message}
\label{TToutput-1}
\rule{6.5in}{.010in}
\end{figure*}

This transformational process differs from previous text planning
efforts in several ways.  First, 
TraumaGEN's eventual message must achieve several top-level
communicative goals as a unit.  This differs from traditional
text planners that must satisfy a single rhetorical goal such
as relating the temporal sequence of events in which a particular
ship is a part (Hovy, 1991) or achieve a single intentional
goal such as getting the user to make a particular change in
a Lisp program (Moore and Paris, 1993). In such cases, the text planner
can construct a plan top-down from the single 
goal and include those propositions that fit into a coherent
piece of text and contribute to achieving the top-level goal.
Although Wolz (1990) developed a system for generating text
satisfying dual, but related, discourse goals such as responding and
enriching, her system focused on eliminating obvious or
redundant information, not on producing integrated messages from
individual, possibly conflicting, critiques.
Similarly, the WISHFUL system (Zukerman and McConachy, 1995) includes 
an optimization stage during which it chooses the optimal way to
achieve a set of related communicative goals; however, the system can
choose to eliminate propositions and it does not have to
deal with conflict within the information to be conveyed.

Second,
the means for achieving
each of the individual goals has already been identified by other
modules.  Thus TraumaGEN is not responsible for
identifying the content of the message but must instead determine how to
realize an effective overall message from the set of individual 
critiques.  We note that this problem is likely to
arise elsewhere as sophisticated systems distribute their
processing across individual modules, each of which may need to
communicate with the user.  

Although natural language has been used in other health care systems
such as MIGRAINE (Buchanan et al., 1995) and 
Piglet (Binsted, Cawsey, and Jones, 1995),
their applications have not required that they combine several independent
but inter-related
text plans into a single integrated message.
The work most closely related to
ours is HealthDoc (Hirst and DiMarco, 1996; Wanner and Hovy, 1996); 
however, HealthDoc (currently under
development) focuses on editing sentences selected from a master
text, such as by inserting pronouns or by deleting references to
propositions that do not appear earlier in the selected text. 

\subsection{Constructing Effective Message Units}
The nature of trauma management and our observations of
communication in the emergency room trauma bay suggested several 
features that should influence the generation process:
\begin{itemize}
\item
\underline{\em Purpose}:  Since the
purpose of messages is to support decision-making, the system's
recommendations should continue to be
organized in terms of relevant domain goals, so that the physician can
evaluate the system's recommendations and decide whether to
adopt them.
\item
\underline{\em Situation}:
Since the emergency room is a chaotic setting for time-critical decisions,
the messages must be succinct, unambiguous, and easily assimilated.
\item
\underline{\em Social role}:
Since the system's social role on the medical team is that of an
expert consultant to the physician who retains responsibility for the
quality of patient care, it must recognize that
the physician can ignore its recommendations.  This differs
from other scenarios, such as tutoring, where the system is the sole
arbiter of correct behavior.
\end{itemize}
Our transformational rules take these factors into account.

\subsubsection{Informational Overlap}
One prevalent problem in TraumaTIQ's output is {\em informational
overlap}  --- actions often appear 
in several different warnings and
thus the message as a whole appears repetitious.  For example,
the upper half of
Figure~\ref{TToutput-1} presents a set of five individual warning
messages generated by TraumaTIQ at one point in a case.
Each warns about the omission of warranted
actions; the fifth one is realized differently due to the lesser estimated
{\em disutility} of the identified error.
While it seems obvious that this set of comments should be combined
into a more coherent message, it is much less clear which of many ways
to effect the combination.
Our approach for merging critiques is
motivated by four often conflicting goals:
1)~group by relevant treatment goals,
2)~avoid repeated mention of the same actions, since this can
erroneously suggest multiple instances of the action,
3)~produce concise messages, and
4)~produce few, rather than many, individual messages.

\begin{figure}
\centerline{\epsfxsize=3.5in\epsffile{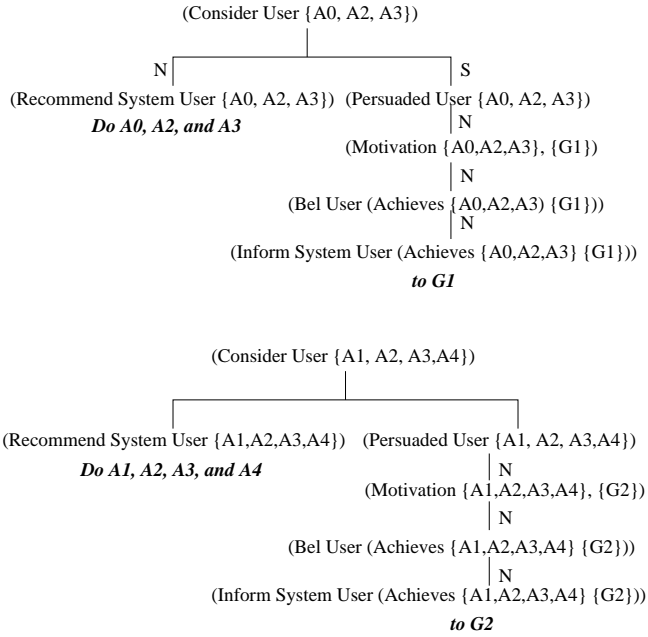}}
\caption{Two Individual Text Plans}
\label{TextPlans}
\rule{3.2in}{.010in}
\end{figure}

\begin{figure*}
\centerline{\epsfxsize=7.5in\epsffile{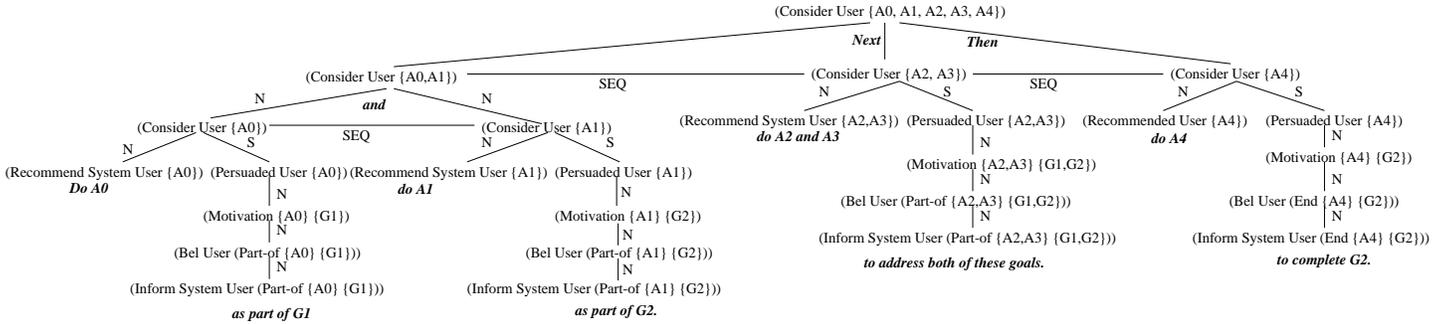}}
\caption{Text Plan for First Merged Message}
\label{CombinedPlan}
\rule{6.5in}{.010in}
\end{figure*}

Rhetorical structure 
theory (Mann and Thompson, 1987)  posits that a coherent text plan
consists of segments related to one another by rhetorical relations such
as motivation or background.
To address the problem of informational overlap, we found that it
was necessary to draw on the multi-nuclear {\em Sequence} relation  
of RST. 
We posited that separate plans for similar
communicative goals involving sets of recommended actions
in the original messages (such as the two plans in
Figure~\ref{TextPlans}) could be reorganized as a sequence of communicative
goals in a single plan, 
with the recommended actions distributed over
the sequentially related goals (as in the plan in Figure~\ref{CombinedPlan}),
as long as the new plan captures the relationships between
recommended actions and their motivations given in the original plans.
Reorganizing messages into sequentially related 
goals allowed us to construct merged messages that exploit informational
overlap in the individual messages yet still achieve the
goals of the original messages.

Thus our first rule, {\em Combine-Similar-Intentions}, looks for overlap
among the components of 
individual messages that have a similar communicative goal
(such as a goal of getting the physician to perform some
omitted actions) and
evaluates the resultant message using a metric that weighs
1)~the number of segments that a goal is distributed over
in the resultant message,
2)~the reduction in repetition of actions,
3)~whether goals must be repeated, and
4)~the number of individual critiques that are merged.

The first term in our metric measures how much the original
messages were reorganized.  Since the original messages represent
an organization in terms of treatment goals,  
the more segments
comprising a merged message, the less the message is organized
in terms of how to address a relevant goal and the more it 
resembles an action recipe.
The merged message in Figure~\ref{TToutput-1} consists of
two segments, each realized as a sentence.  One consequence
is that the goal of treating the compound rib fracture is now
distributed across both segments.    
We arbitrarily limit combined messages
to three segments in order to maintain a goal-oriented organization,
as dictated by the {\em purpose} of our messages.

The next three components of our metric measure how well a merge
achieves  concise, unambiguous, and easily assimilated messages
(as required by the {\em situation} in which the messages will be
received).
The reduction in repetition of actions contributes both to concise
messages and decreasing ambiguity; however, achieving this often
requires the repetition of goals, which detracts from the succinctness
of the message.  We hypothesize that a few coherent messages will
be more easily assimilated than many individual messages and thus our
metric takes into account the number of individual critiques that are
merged into the resultant message.

Figure~\ref{TextPlans} illustrates the text plans underlying the
individual critiques:
\begin{center}
\begin{tabular}{p{3.2in}}
Caution: do $<A_0>$, $<A_2>$, and $<A_3>$ to \mbox{$<G_1>$}.
\\
Caution: do $<A_1>$, $<A_2>$, $<A_3>$, and $<A_4>$ to $<G_2>$.
\end{tabular}
\end{center}
while Figure~\ref{CombinedPlan} illustrates the text plan produced by
TraumaGEN for the merged message
\begin{center}
\begin{tabular}{p{3.2in}}
Do $<A_0>$ as part of $<G_1>$ and
$<A_1>$ as part of $<G_2>$.
Next do $<A_2>$ and $<A_3>$ to address both of these goals.
Then do $<A_4>$ to complete $<G_2>$.
\end{tabular}
\end{center}
For reasons of efficiency and real-time response, TraumaGEN applies its
transformational rules 
directly to the logical form of the critiques
produced by TraumaTIQ.
An example of an actual merged message produced by TraumaGEN is shown in
Figure~\ref{TToutput-1}.
  
\subsubsection{Trailing Comments}

When several critiques are merged into a single message, the message may
refer to actions that are also part of critiques that did not
participate in the merge.
Once these actions have been introduced in the merged message, 
discourse theory, particularly work on focusing
heuristics (McKeown, 1985), suggest that the other
critiques referencing these
actions be included in the merged message as well.  However, rather
than restructure the result of our merge transformation, we
append them to the end of the message.  Thus we refer to them
as {\em trailing comments}.

Unfortunately, trailing comments have the potential to erroneously
suggest new instances of actions.  Our solution to this problem is
to (1) make the previously focused action the subject of the sentence,
reflecting its {\em given} status in the discourse, and (2) utilize cue
words to call additional attention to its occurrence earlier in the
message and to the new information being conveyed.  
Thus the first trailing comment is introduced with the
cue  word {\em moreover} since this cue word carries the implication of saying
more about something already discussed, and the cue word
{\em also} is used to introduce the additional information.
In the following critique, for example,  
the final sentence (underlined for exposition) contains
a trailing comment:
\begin{center}
\begin{tabular}{p{3.2in}}
{\em Check for distended neck veins and decreased breath sounds
to assess the possibility of a left tension pneumothorax and 
a pericardial tamponade.  Then check for muffled 
heart sounds and continued shock to complete assessing the possibility
of a pericardial tamponade.}
\underline{\em Moreover, checking}
\underline{\em for muffled
heart sounds is also indicated to assess the} 
\underline{\em  possibility of a pericardial injury.} 
\end{tabular}
\end{center}
If there is a second trailing 
comment,
it is then introduced with the cue phrase {\em in addition} since it
suggests presenting new information of a similar nature, in this
case another reference to a previously introduced action. 

A trailing comment may need to refer to other actions in addition to the one
previously focused on.
We accomplish this by subordinating those actions in a phrase introduced
by the cue phrase {\em along with}, in a sentence in which the previously
focused action is the subject.  For example, in Figure~\ref{TToutput-1},
the fourth critique is not included in the merged message but
includes an action
({\em doing a laparotomy}) that is part of the merged message.  Thus
the fourth critique is realized as a trailing comment at
the end of the merged message:
\begin{center}
\begin{tabular}{p{3.2in}}
{\em Moreover, doing the laparotomy
is also indicated, along with repairing the left
diaphragm, to treat the lacerated left diaphragm.} 
\end{tabular}
\end{center}
%

\subsubsection{Revising Interacting Critiques}
\begin{figure}
\begin{center}
\begin{tabular} {p{3.2in}}
{\em Performing local visual exploration of all abdominal
wounds is preferred over doing a peritoneal lavage for ruling out a
suspicious abdominal wall injury.} \\

\\{\em Please remember to check for laparotomy scars before 
you do a peritoneal lavage.}
\end{tabular}
\end{center}
\caption{Two Conflicting Critiques}
\label{Conflict}
\rule{3.2in}{.010in}
\end{figure}
In TraumaTIQ's original output, we noticed that one
critique could detract from another critique, although each was both
justified and coherent in isolation.
Consider, for
example, the two critiques shown in Figure~\ref{Conflict}.
The first cautions the physician that a procedure other than
the just-ordered peritoneal lavage is
the recommended procedure in this instance, although
the disparity is not critical. The second reflects the fact that a
peritoneal lavage should not be done on someone with abdominal
scarring.  Since TraumAID does not yet have any information 
about the presence of abdominal scarring in this patient,
the critique reminds the physician of the need to first check for it.

However, together the two critiques appear incoherent.  This can be
more pronounced when the two critiques are separated by
other comments, since the second critique gives the impression that
a peritoneal lavage will be performed.
In some situations, such as tutoring, it might be appropriate to discard
the second critique.  However, in
real-time decision-support, this may be inappropriate since it
presumes that the system is the sole arbiter of high-quality performance whose
advice must be followed.
\begin{figure*}
\centerline{\epsfxsize=6.5in\epsffile{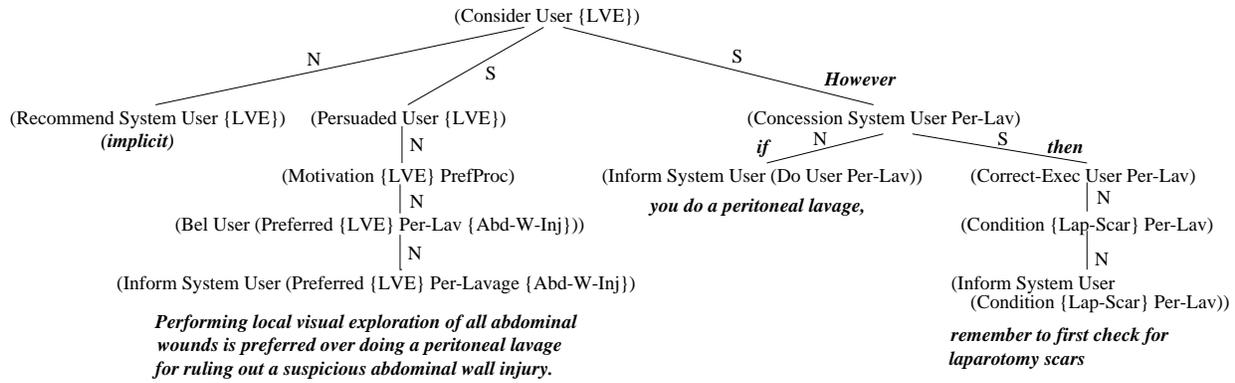}}
\caption{Text Plan for Revised Message}
\label{RevisedPlan}
\rule{6.5in}{.010in}
\end{figure*}

Our solution to this problem is to allow revision rules that are triggered
when two critiques have a potential conflict.
In such cases, the two critiques are merged into a single message,
and the conflicting critique is revised.  For example, our rule
{\em Revise-Conflict} is triggered whenever a critique whose
goal is that an action be properly scheduled occurs with a critique
whose intention is that the action be replaced with one more highly
preferred.
{\em Revise-Conflict} merges the two critiques into a single
message, where the merged message reflects a 
{\em concession} that the original action might still be executed
and the scheduling critique is revised so that
it is conditionally dependent on the original action being done. 
For example, in the case of
the critiques in Figure~\ref{Conflict}, the two critiques would be
revised and merged into the single message
\begin{center}
\begin{tabular}{p{3.2in}}
{\em Performing local visual exploration of all abdominal wounds
is preferred over doing a peritoneal lavage for ruling out a 
suspicious abdominal wall injury.  However, if you do a peritoneal
lavage, then remember to first check for laparotomy scars.}
\end{tabular}
\end{center}
Figure~\ref{RevisedPlan} illustrates the text plan underlying the
revised message.
Note that the new message still recommends the better
procedure, but leaves the final choice with the physician
who is responsible for quality patient care.  
On the other hand, if the second critique in Figure~\ref{Conflict}
appeared by itself, no revision of the message would occur.

Another rule, {\em Revise-Interactions}, is triggered
when a critique whose goal is to postpone a dependent action
occurs in conjunction with a critique whose goal is execution of
the action on which the dependency is based.  For example,
Figure~\ref{Dependent} presents two of TraumaTIQ's critiques.
While the two critiques do not conflict, 
the relation between their communicative goals
can be exploited to produce a more concise and coherent message.
{\em Revise-Interactions} establishes a {\em Sequence} relation
between doing the peritoneal lavage and the decision about whether
to do the reassessment,  and produces the message
\begin{center}
\begin{tabular}{p{3.2in}}
{\em Do a peritoneal lavage immediately as part of ruling out
abdominal bleeding.  Use the results of the peritoneal lavage
to decide whether or not to reassess the patient in 6 to 24 hours.}
\end{tabular}
\end{center}

\subsection{Other Influences on Effective Messages}
\subsubsection{The Role of Focus}

\begin{figure}
\begin{center}
\begin{tabular}{p{3.2in}}
{\em Caution: do a peritoneal lavage immediately as part of ruling out
abdominal bleeding.} \\ \\
{\em Do not reassess the patient in 6 to 24 hours until after doing
a peritoneal lavage.  The outcome of the latter may affect the need to
do the former.}
\end{tabular}
\end{center}
\caption{Two Dependent Critiques}
\label{Dependent}
\rule{3.2in}{.010in}
\end{figure}
Focus (Grosz, 1977; McKeown, 1985)
has been the objective of much discourse research, and it plays several roles 
in our generation of messages.
As noted earlier, trailing comments capture communicative goals
that relate to previously mentioned actions, and cue words are
used to shift focus back to the earlier actions.  In addition,
if there is more than one trailing comment, they appear in order
of the most recently introduced action, thus representing successive
pops of a focus stack.

Focus also affects the way in which some communicative goals are realized
in messages.
For example, if a goal of getting the user to recognize several scheduling
constraints is the sole content of a message, it would be realized
with the subordinate clause first to call attention to the ordering
constraint, as in the following:
\begin{center}
\begin{tabular}{p{3.2in}}
\underline{\em Before getting the urinalysis,} 
{\em insert the left chest tube
and get the post chest tube x-ray because they 
have a higher priority.}
\end{tabular}
\end{center}
However, if the physician has omitted some of the actions and the
scheduling constraint is incorporated into  the 
text plan for getting the physician
to do the omitted actions, then focus considerations dictate that the
main clause appear first since it continues the actions in focus.
The following is such an example produced by our system:
\begin{center}
\begin{tabular}{p{3.2in}}
{\em Check for medication allergies, give antibiotics, set up the 
auto-transfuser for the left chest tube, insert a left chest tube, and
get a post chest tube x-ray  to treat the
simple left hemothorax.}  \underline{\em Insert the}
\underline{\em left chest tube and get the post chest tube x-ray before} 
\underline{\em doing the peritoneal lavage because they have a higher}
\underline{\em priority.}
\end{tabular}
\end{center}

\subsubsection{Definite Versus Indefinite References}

Critiques or any other message from the system should be
phrased in terms of what is {\em shared knowledge} in the emergency
room. We equate shared knowledge with the current state of the
case, as it has been entered into the computer-based medical  record (CBMR).
When a procedure is ordered, it thus becomes part of this shared knowledge.
Consequently, we use definite articles to refer both to procedures
and actions already
introduced into the treatment plan by 
one of the system's messages and to entities introduced via
the scribe nurse's entry of a procedure or action into the CBMR.
For example, even though a peritoneal lavage does not appear in any of
the system's earlier messages, a message about a 
related scheduling precondition
will be realized as:
\begin{center}
\begin{tabular}{p{3.2in}}
{\em Please remember to check for laparotomy scars before you
do \underline{the peritoneal lavage}.}
\end{tabular}
\end{center}

However, the system may disagree with the physician about whether a procedure
is appropriate.  Since the use of the definite article suggests an
action's acceptance into the treatment plan,
we use indefinite expressions when referring to procedures
about which there is conflict.  For example, if the physician has ordered a
peritoneal lavage and the system believes that the need for it is
dependent on the results of a chest x-ray, the system
would generate the message
\begin{center}
\begin{tabular}{p{3.2in}}
{\em Do not do \underline{a peritoneal lavage} until after getting a
chest x-ray since
the outcome of the latter may affect the need to do the former.}
\end{tabular}
\end{center}

\section{Evaluation}

While the final evaluation of TraumaGEN's effectiveness can only be measured
by deploying it in a trauma bay and evaluating the degree to which its
messages change physician behavior, preliminary evaluation can be used
to determine its benefits and
limitations and to identify where
further work is needed.  
We ran TraumaGEN on 
48 collected cases of actual trauma care under a scenario
in which critiques were produced after each physician order.

We compared the critiques generated by TraumaTIQ alone  with the
messages produced when it was augmented
with TraumaGEN.  We found that TraumaGEN reduced the number of messages,
resulted in more concise messages (measured in terms of the number of
noun phrases in a message), and required fewer shifts in focus
to assimilate the messages.

To evaluate coherence and quality of the messages, we asked a human
subject not affiliated with our project to evaluate the new messages
with respect to the original ones.
The subject
was given the messages produced by TraumaTIQ and TraumaGEN for  a dozen
cases.  
In ten of the twelve cases, the human subject preferred
TraumaGEN's messages; in eight of these cases, the preference was
very strong while in the other two cases it was moderate.
In the single instance in which the subject preferred the original
messages produced by TraumaTIQ, the preference was based on the 
English translation of two goals; the subject found the phrasing
confusing when the messages were combined (since the two goals had
very similar translations) but not confusing when the messages
were separated.  
The subject's comments indicated that his preferences for TraumaGEN's
messages were generally based on reduction of repetition, merging
of related messages, and elimination of conflict. 

\section{Conclusion}

This paper has presented our message planner, TraumaGEN, 
that draws on rhetorical structure
and discourse theory to produce
integrated messages from individual critiques each of which
is designed to achieve its own communicative goal.  The need to
construct coherent text from multiple individual text plans
is a problem that will increasingly face natural language systems
as sophisticated systems distribute their processing across
individual modules each of which may need to communicate with the
user. TraumaGEN  takes into account knowledge about the {\em purpose} of
the messages, the
{\em situation} in which the messages will be received, and 
the {\em social role} of the system.
Preliminary evaluation of TraumaGEN indicates that 
it successfully constructs coherent
integrated messages from individual critiques 
and that the resultant messages are
comprehensible and a significant improvement over the original critiques.

\section{Acknowledgments}

This work was supported by the National Library of Medicine
under grant R01-LM-05764-01.  We would like to thank Bonnie Webber
and John Clarke for their many helpful suggestions on this work and
their comments on this paper.

\bibliography{/usa/carberry/cargroup/bibfile}

\newcommand{\etalchar}[1]{$^{#1}$}
\begin{thebibliography}{BMF{\etalchar{+}}95}

\bibitem[BCJ95]{Binsted95}
K.~Binsted, A.~Cawsey, and R.B. Jones (1995).
\newblock Generating personalized information using the medical record.
\newblock {\em Proceedings of AIME},  29--41.

\bibitem[BMF{\etalchar{+}}95]{Buchanan95}
B.~Buchanan, J.~Moore, D.~Forsythe, G.~Carenini, S.~Ohlsson, and G.~Banks (1995).
\newblock An intelligent interactive system for delivering individualized
  information to patients.
\newblock {\em Artificial Intelligence in Medicine},  117--154.

\bibitem[Gro77]{Grosz77a}
B. Grosz (1977).
\newblock {The representation and use of focus in a system for understanding
  dialogs}.
\newblock In {\em Proceedings of the International Joint Conference on
  Artificial Intelligence},  67--76.

\bibitem[GW96]{Gertner96}
A.~Gertner and B.~L. Webber (1996).
\newblock {A bias towards relevance: Recognizing plans where goal minimization
  fails}.
\newblock In {\em Proceedings of the Thirteenth National Conference on
  Artificial Intelligence}, 1133--1138.

\bibitem[HD96]{Hirst96}
G.~Hirst and C.~DiMarco (1996).
\newblock Automatic customization of health-education brochures for individual
  patients.
\newblock In {\em Proceedings of the Conference on Information Technology in
  Community Health}.

\bibitem[Hov91]{Hovy91a}
E. Hovy (1991).
\newblock Approaches to the planning of coherent text.
\newblock In {\em Natural Language Generation in Artificial Intelligence and
  Computational Linguistics},  153--198. Kluwer.

\bibitem[McK85]{McKeown85}
K. McKeown (1985).
\newblock {\em {Text Generation}}.
\newblock Cambridge: Cambridge University Press.

\bibitem[MP93]{Moore93b}
J. Moore and C. Paris  (1993).
\newblock Planning text for advisory dialogues: Capturing intentional and
  rhetorical information.
\newblock {\em Computational Linguistics}, 19(4):651--695.

\bibitem[MT87]{mannthompson87}
W. Mann and S. Thompson (1987).
\newblock {Rhetorical Structure Theory: A theory of text organization}.
\newblock Technical Report ISI/RS-87-190, ISI/USC.

\bibitem[WH96]{Wanner96}
L. Wanner and E. Hovy (1996).
\newblock The HealthDoc sentence planner.
\newblock In {\em Proceedings of the International Workshop on Natural Language
  Generation},  1--10.

\bibitem[Wol90]{Wolz90}
U. Wolz (1990).
\newblock An object oriented approach to content planning for text generation.
\newblock In {\em Proceedings of the International Workshop on Natural Language
  Generation},  95--104.

\bibitem[WRC92]{Webber92}
B. Webber, R. Rymon, and J. Clarke.
\newblock Flexible support for trauma management through goal-directed
  reasoning and planning.
\newblock {\em Artificial Intelligence in Medicine}, 4:145--163, 1992.

\bibitem[ZM95]{Zukerman95a}
I. Zukerman and R. McConachy (1995).
\newblock Generating discourse across several user models: Maximizing belief
  while avoiding boredom and overload.
\newblock In {\em Proceedings of the International Joint Conference on
  Artificial Intelligence},  1251--1257.

\end{thebibliography}
\end{document}